%% file: ijcai22.tex
\documentclass{article}
\pdfoutput=1
\usepackage{ijcai22}

\usepackage{times}
\usepackage{soul}
\usepackage{url}
\usepackage[hidelinks]{hyperref}
\usepackage[utf8]{inputenc}
\usepackage[small]{caption}
\usepackage{graphicx}
\usepackage{amsmath}
\usepackage{amssymb}
\usepackage{amsthm}
\usepackage{booktabs}

\usepackage{helvet}
\usepackage{algorithm}
\usepackage{algorithmic}

\usepackage{amsfonts}       
\usepackage{nicefrac}       
\usepackage{microtype}      
\usepackage{xcolor}         
\usepackage{bm}
\usepackage{makecell}
\usepackage{adjustbox}
\usepackage{wrapfig,lipsum}
\usepackage{bibentry}
\usepackage{newfloat}
\usepackage{listings}
\usepackage{bm}

\usepackage{pifont}         
\usepackage[normalem]{ulem}
\usepackage[capitalize]{cleveref}

\title{Real-Time Portrait Stylization on the Edge}

\author{
    Yanyu Li$^{*1}$\and 
    Xuan Shen$^{*1}$\and 
    Geng Yuan\thanks{These authors contributed equally to this work.}$^1$\and
    Jiexiong Guan$^2$\and \\
    Wei Niu$^2$\and
    Hao Tang$^3$\and
    Bin Ren$^2$ \and 
    Yanzhi Wang$^1$
    \affiliations
    $^1$Northeastern University  \\
    $^2$College of William \& Mary  \\
    $^3$CVL, ETH Zurich
    \emails
    \{li.yanyu, shen.xu, yuan.geng, yanz.wang\}@northeastern.edu,\\
    \{jguan, wniu\}@email.wm.edu,
    bren@cs.wm.edu, hao.tang@vision.ee.ethz.ch
}

\begin{document}

\maketitle

\begin{abstract}
In this work we demonstrate real-time portrait stylization, specifically, translating self-portrait into cartoon or anime style on mobile devices. 
We propose a latency-driven differentiable architecture search method, maintaining realistic generative quality.
With our framework, we obtain $10\times$ computation reduction on the generative model and achieve real-time video stylization on off-the-shelf smartphone using mobile GPUs. 
\end{abstract}

\input{sections/1_introduction}
\input{sections/2_background}
\input{sections/3_method}

\input{sections/4_evaluation}

\appendix

\newpage

\section*{Acknowledgments}
This research is partially supported by National Science Foundation CNS-1909172, CCF-1919117, and CCF-2047516 (CAREER). 

\bibliographystyle{named}
\bibliography{ijcai22}

\end{document}

%% file: sections/1_introduction.tex
\section{Introduction}
Thanks to hardware advancement, varieties of AI applications have been enabled on portable smart devices, such as foreground segmentation, face recognition, etc. 
In this work, we investigate portrait stylization which is a popular feature in social media Apps,
transferring self portraits into a desired style, such as cartoon~\cite{andersson2020generative}, anime~\cite{Kim2020U-GAT-IT:,li2021anigan}, etc. 

Portrait stylization can be categorized as classic image-to-image translation, which is often achieved by conditional Generative Adversarial Networks (GANs) \cite{isola2018imagetoimage,CycleGAN2017}.
In GAN training, a generator learns to generate a fake instance and fool the discriminator, while the discriminator takes true and generated images as input and learns to distinguish them. 
Consequently, in the portrait stylization task, the generator is utilized to map a portrait photo to the desired style domain during inference. 
Compared to naive paired supervised training, GANs exhibits stunning generative quality, e.g., sharp and realistic details, diverse features.

However, it still remains challenging to enable real-time face stylization to process videos on a mobile device, and the reason comes two-fold. 
Firstly, typically following an encoder-decoder design, image translation models suffer from high computation complexity, especially on high-resolution images.
Secondly, GAN training is difficult and unstable, suffering from loss divergence and mode collapse. 
As a result, existing compression techniques are difficult to integrate into GAN training and preserve generative quality.

\begin{figure}[t]
    \centering
    \includegraphics[width=0.4\textwidth]{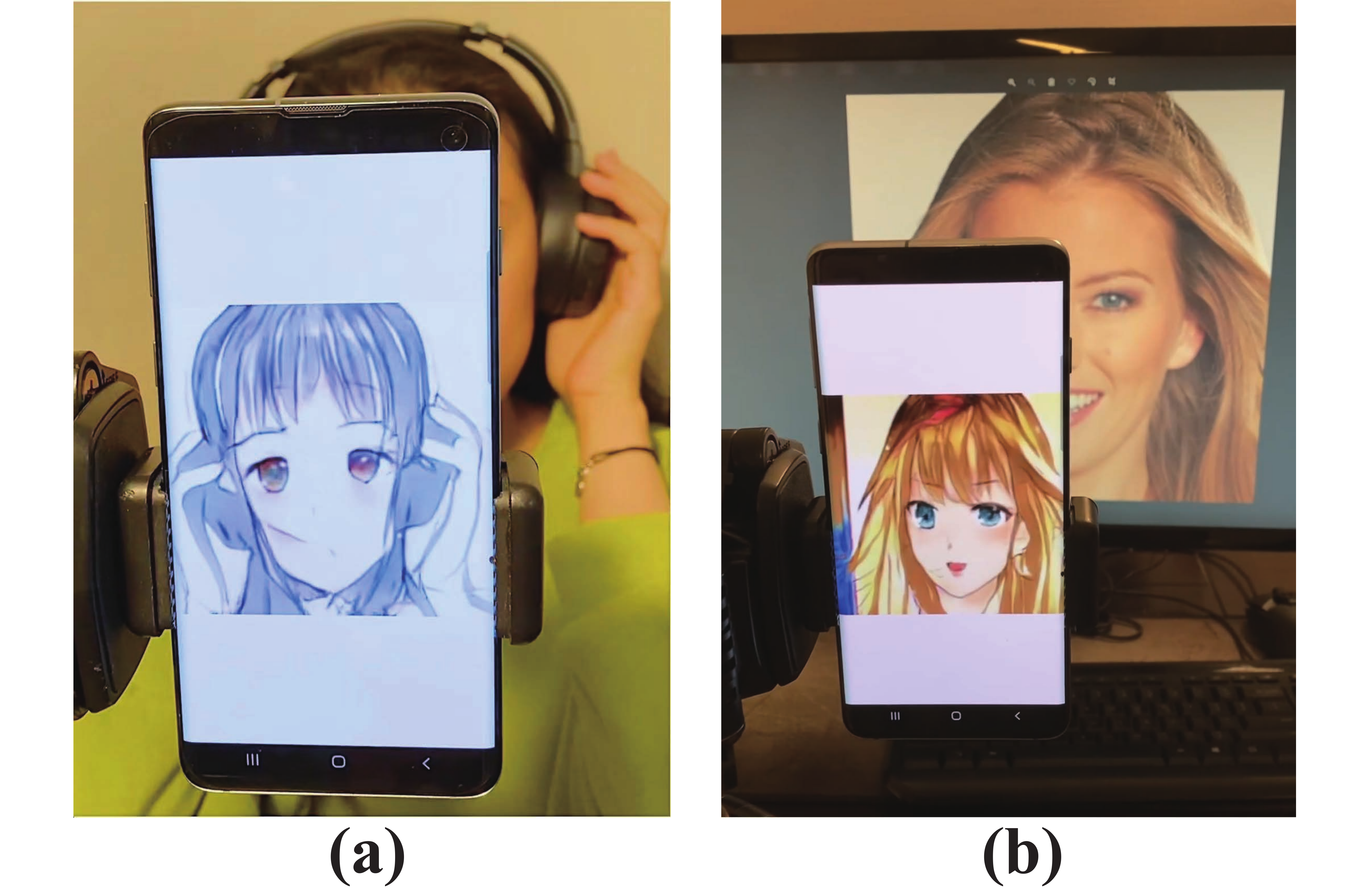}
    \caption{Demonstration of on-mobile real-time stylization for (a) real person and (b) static image. The input video stream is captured by the mobile phone's camera in a real-time fashion. The full demo video is attached in the link
    .
    }
    \label{fig:demo}
\end{figure}

In this work, we propose a compiler-aware differentiable architecture search framework. 
We measure the latency of the building blocks with sufficient configurations (channels, feature sizes), and train a neural network to predict the latency. 
We show that a simple MLP speed model can make accurate predictions. 
In order to search compact architectures, we integrate learnable parameters in the GAN generator and regularize them by speed constraints to reduce the model width and depth. 
With the speed model that maps architecture parameters to latency, the speed penalty is differentiable so that we can perform search with end-to-end training. 
Plus, different from prior work, we do not select preserved/eliminated model weights by magnitude. 
Instead, we apply straight through estimator on architecture parameters to sparsify them into $\{0, 1\}$. 
The benefit is two-fold. 
The remained weights are represented by $1$s so that we can easily predict the latency of a certain state. 
In addition, the gradients of pruned weights are completely zeroed out so that their functionalities are preserved.
As a result, pruned weights are always ready to be reverted back to contribute to accuracy during exploration. 
This is especially important in GAN search because of the unstable training process. 

Overall, our contributions include:
\begin{itemize}
    \item We develop a latency-driven differentiable architecture search for GANs. Our sophisticated pipeline addresses the difficulty in GAN searching, achieves unprecedented compression rate while preserving generative quality. 
    \item To the best of our knowledge, we are the first to achieve real-time portrait stylization on mobile phones. Mobile demos are attached in the link\footnote{The link for the demo video: \url{https://youtu.be/2SFAVwaymvQ}
    }. 
\end{itemize}

%% file: sections/2_background.tex
\section{Background}

\paragraph{Image-to-Image Style Transfer.}
\cite{isola2018imagetoimage,liu2018unsupervised,huang2018munit,DRIT,liu2019few,park2020cut,Kim2020U-GAT-IT:,li2021anigan,chong2021gans,tang2019multi} aim to translate a image from the source domain to match certain styles in the target domain, while preserving semantics of the origin image. 
Early works train the generative model with paired data \cite{isola2018imagetoimage}, however they cannot be applied to more common unpaired datasets.
Later work \cite{CycleGAN2017} proposed a cycle consistency loss to train on unpaired data domains, inspiring lots of successor research on image stylization \cite{park2020cut,Kim2020U-GAT-IT:,li2021anigan,chong2021gans,tang2021attentiongan}. 

As for the domain of cartoon or anime, UGATIT \cite{Kim2020U-GAT-IT:} released the selfie2anime benchmark and develop an adaptive mixture of instance and layer normalization. 
AniGAN \cite{li2021anigan} released a larger scale face2anime dataset and studied style-guided anime translation. 
GNR \cite{chong2021gans} refine the content and style to produce controllable and diverse synthesis.

\paragraph{Compressing GANs.}
Because of wide applications, the compression of GAN has drawn research attention \cite{fu2020autogandistiller,wang2020ganslimming,li2020gan,chen2021gans,jin2021teachers}. 
Recent work \cite{li2020gan} proposed to incorporate neural architecture search (NAS) and feature level knowledge distillation (KD).
\cite{jin2021teachers} integrated an Inception-like residual block and performed self-distillation.

%% file: sections/3_method.tex
\section{Compiler-Aware Architecture Search}

\begin{figure*}[ht]
    \centering
    \includegraphics[width=0.85\textwidth]{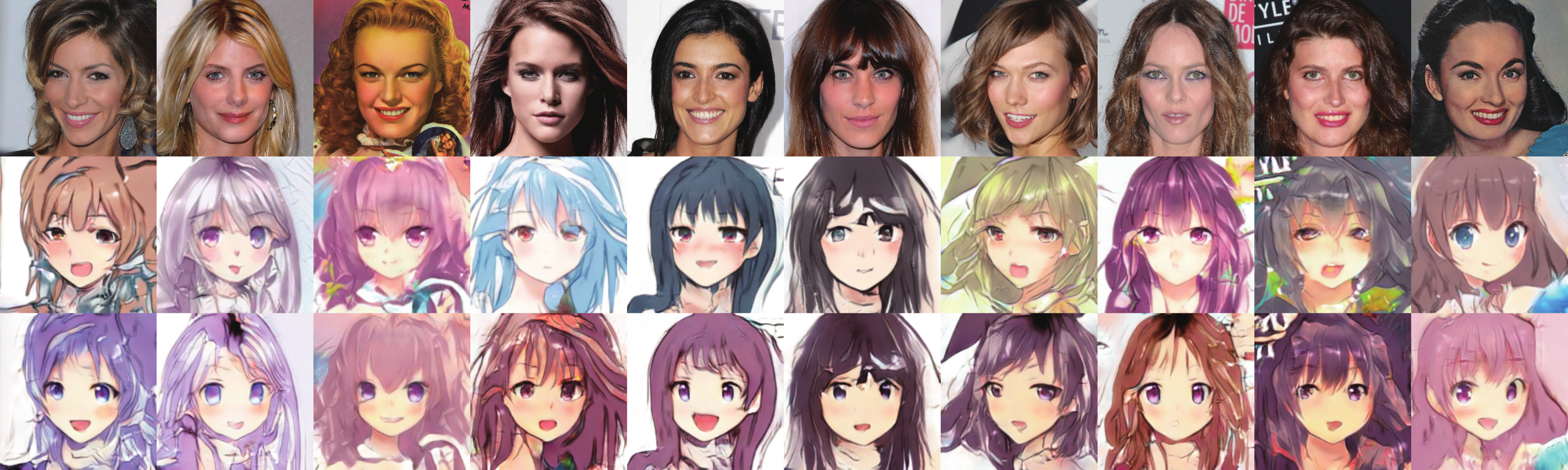}
    \caption{Stylization results on face2anime dataset. Top: source image. Middle: styles generated by baseline UGATIT model with 57.1 GMACs. Bottom: our searched model under 5.56 GMACs. }
    \label{fig:anime}
\end{figure*}

\begin{figure}[ht]
    \centering
    \includegraphics[width=0.4\textwidth]{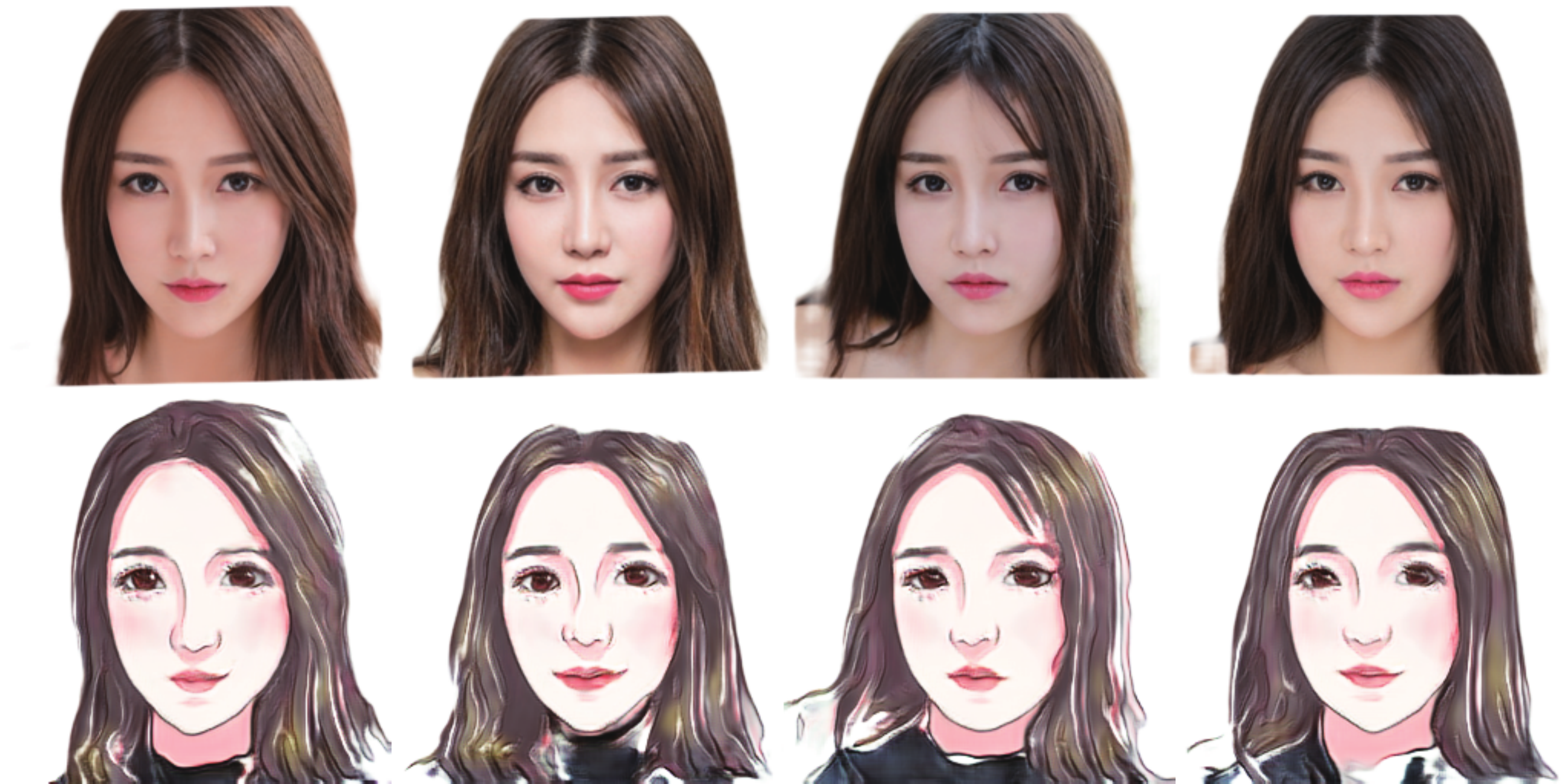}
    \caption{Stylization results on photo2cartoon dataset released (partly) by MiniVision. Top: source image. Note that the source domain data (Asian women) comes from generated fake data by StyleGAN2. Bottom: our searched model under 1.34 GMACs. }
    \label{fig:cartoon}
\end{figure}

\subsection{GAN Basics}
We follow CycleGAN \cite{CycleGAN2017} and UGATIT \cite{Kim2020U-GAT-IT:} paradigm to develop our architecture search for stylization. 
According to CycleGAN, we learn mapping functions between two unpaired domain $X{=}\{x_i\}_{i=1}^N$ and $Y{=}\{y_j\}_{j=1}^M$. 
There are two generators in inverse direction $G{:} X {\xrightarrow[]{}} Y$ and $F{:} Y {\xrightarrow[]{}} X$, as well as two discriminators $D_X$ and $D_Y$. 
Note that we refer $G$ as the generator mapping portraits to stylized images, which is the only required model during inference. 
\paragraph{Adversarial Loss.} We match the distribution of translated instances with the target domain as follows.
\begin{equation}
\small
    L_{gan}^X = \mathbb{E}_{y \sim Y} [D_Y^2(y)] + \mathbb{E}_{x \sim X} [(1-D_Y(G(x)))^2].
\end{equation}

\paragraph{Cycle Consistency Loss.} To minimize reconstruction error,
\begin{equation}
\small
    L_{cyc} = \mathbb{E}_{x \sim X} [|F(G(x)) - x|_1] + \mathbb{E}_{y \sim Y} [|G(F(y))-y|_1]. 
\end{equation}

Note that we also incorporate the identity loss $L_{id}$ and CAM loss $L_{CAM}$ as proposed in UGATIT, for simplicity we skip detailed formulations here and please refer to \cite{Kim2020U-GAT-IT:}. 

Our overall GAN objective is: 
\begin{equation}
\small
    L =  \lambda_1 L_{gan}^X + \lambda_1 L_{gan}^Y + \lambda_2 L_{cyc} + \lambda_3 L_{id} + \lambda_4 L_{CAM},
\end{equation}
where $\lambda_1{=}1, \lambda_2{=}10, \lambda_3{=}10, \lambda_4{=}1000$ are the hyperparameters to control each loss.

\subsection{Layerwise Width Search}

Width search is performed for each CONV layer. 
We choose the supernet from \cite{CycleGAN2017,Kim2020U-GAT-IT:}, which is a commonly employed generator. 
In order to create a learnable binary mask, we insert a depth-wise $1{\times} 1$ CONV  layer following each CONV layer to be pruned, as shown below, 
{\small
\begin{equation}
  \bm  a_l^n =  \bm v_l^n \odot (\bm w_l^n \odot \bm a_{l-1}^n),
\end{equation}}%
where $\odot$ denotes the convolution operation. $\bm w_l^n {\in} R^{o\times i\times k\times k}$ is the weight parameters in the $l^{th}$ CONV layer of the $n^{th}$ block, with $o$ output channels, $i$ input channels, and  kernels of size $k{\times} k $. 
$\bm a_l^n {\in} R^{B\times o\times s\times s'}$ represents the output features of $l^{th}$ layer (with the trainable mask), with $o$ channels and  $s{\times} s'$ feature size. $B$ denotes the batch size. 
$\bm v_l^n {\in} R^{o\times 1\times 1\times 1}$ is the corresponding weights of the depth-wise CONV layer (i.e., the mask layer).

Larger elements of $\bm m_l^n$  mean that the corresponding channels should be preserved while smaller elements indicate that the corresponding channels should be pruned. 
Formally, we use a threshold $t$ to convert $\bm m_l^n$ into a binary mask as below,
{\small
\begin{equation}
 \bm b_l^n = 
    \begin{cases}
    1, \bm v_l^n > t. \\
    0, \bm v_l^n \leq t.
    \end{cases} \text{(element-wise)}, 
\end{equation}}%
where $\bm b_l^n {\in} \{0, 1\}^{o\times 1\times 1\times 1}$ is the binarized $\bm v_l^n$.
Typically, we initialize $\bm v_l^n$  with 1, and the adjustable $t$ is  set to 0.5. 
In order to make the mask differentiable to enable backpropagation, we utilize Straight Through Estimator (STE) \cite{bengio2013setimating,chang2020mix} as shown below, 
{\small
\begin{equation}
\frac{\partial \mathcal L}{\partial \bm b_l^n} = \frac{\partial \mathcal L}{\partial \bm v_l^n}.
\end{equation}}%
Our trainable binary mask has the following advantages:  
(i) The mask can be trained along with the  network parameters via  gradient descent, thus saving search cost compared to NAS methods \cite{zoph2016neural,zhong2018practical}. 
(ii) Different from previous methods \cite{han2015learning,yu2017compressing,he2017channel,guan2020dais}, which determine the pruning  according to the  parameter magnitudes,  we decouple the parameter magnitudes of CONV or BN layer from pruning, and utilize independent mask layers, thus the remained parameters are not harmed.  
(iii) The discrete values can directly provide the width information for each CONV layer, which is  compatible with speed prediction.

\subsection{Length Search by Block}

Note that although per-layer width search may also converge to zero width, which eliminates the entire block, we find that there are usually a few channels left in each block
preventing us to remove the entire block. 
Plus, the pruning indicator for each layer cannot represent the latency reduction of entirely pruned blocks. 
Thus it is necessary to perform length search separately from width search.

We construct two paths in each customized residual block, one is the masked convolution block and the other is skip connection. 
In the aggregation layer of the dual paths, we integrate binarized variables $\beta_1$ and $\beta_2$, then the forward computation can be represented as follows, 
\begin{equation}
\small
\bm {a}^n  = \beta_1 \cdot \bm a^{n-1}  + \beta_2 \cdot \bm a_L^n.
\end{equation}

The aggregation layer contains two trainable parameters $\alpha_1$ and $\alpha_2$, shares similar STE recipe with width parameter. In the forward pass, it selects the skip path or the masked convolution path based on the relative relationship of $\alpha_1$ and $\alpha_2$, 
{\small
\begin{align}
  \bm  \beta_1 = 0 \   \text{and}  \ \beta_2 =1,  \text{if} \ \alpha_1 \le \alpha_2.  \\  
    \bm  \beta_1 = 1 \  \text{and} \ \beta_2 =0,   \text{if} \  \alpha_1 > \alpha_2.  
\end{align}}%

\subsection{Speed Prediction with Speed Model} \label{sec: speed_model_framework}
We take inference speed on mobile GPUs to constrain the optimization.  
A DNN-based speed model is adopted to predict the inference speed of the block based on its architecture configurations. 
Then the final predicted latency is accumulated by the aggregated blocks so that we can compute the latency loss $L_{latency}$ and integrate it into the searching pipeline. 
\begin{equation}
    L_{search} = L + L_{latency}(\bm v, \alpha).
\end{equation}

The trained speed model is accurate in predicting the speed of different layer widths in the block (with 5\% error at most).

\begin{table}[] 
\small
\centering
\begin{tabular}{ccc}
\toprule
Face2anime & FID $\downarrow$   & MACs ($\times 10^9$) \\
\hline
CycleGAN~\cite{CycleGAN2017}   & 50.09 & 56.8  \\
UGATIT~\cite{Kim2020U-GAT-IT:}     & 42.84 & 57.1  \\
MUNIT~\cite{huang2018munit}      & 43.75 & 77.3  \\
FUNIT~\cite{liu2019few}      & 56.81 & -     \\
DRIT~\cite{DRIT}       & 70.59 & -     \\
AniGAN~\cite{li2021anigan}     & 38.45 & -     \\
\hline
Ours       & 57.32 & 5.56  \\
\bottomrule
\end{tabular}
\caption{Quantitative evaluation of our searched stylization model on Face2anime dataset. Our model perform inference on video at 18 FPS on SAMSUNG Galaxy S10 smartphone using mobile GPU.}
\label{tab_quality}
\end{table}

%% file: sections/4_evaluation.tex
\section{Experiments and Demonstration}

We conduct experiments on Face2anime dataset published by \cite{li2021anigan}. 
Face2anime consists of 17,796 images. 
We set input size to $256{\times} 256$. 
Further, we also demonstrate realistic quality on cartoon stylization released by MiniVision, along with Asian women face training data generated by \cite{Karras2019stylegan2}. 

\paragraph{Experiment Setups.}
As for the supernet, we search from \cite{Kim2020U-GAT-IT:}. 
Learning rate is set to $2{\times} 10^{-4}$ for both generator and discriminator, with Adam optimizer. 
Leaning rate is fixed for the first 30k iterations and then linearly decayed to zero in another 30k iterations.

\paragraph{Performance Evaluation.}
We quantitatively compare our searched stylization model with representative works in the quality of created images (FID) and computation costs (MACs). As shown in Table~\ref{tab_quality}, our generative model 
achieves 10$\times$ computation reduction and preserves generative quality.
Thanks to the significant computation reduction, we achieve high quality real-time stylization on mobile. 

\paragraph{Results Visualization and On-mobile Demonstration.}
Figure~\ref{fig:anime} shows the visualization of the created images on face2anime test dataset and the comparison of our 5.56 GMACs model with the baseline UGATIT model with 57.1 GMACs.
Figure~\ref{fig:cartoon} shows the created images using photo2cartoon test dataset. 
With simpler styles, our search method further reduce the model size to 1.34 GMACs.
We also demonstrate our method on the mobile device, as shown in Figure~\ref{fig:demo}. 
The full demo video is available in the link
.

%% file: ijcai22.bbl
\begin{thebibliography}{}

\bibitem[\protect\citeauthoryear{Andersson and
  Arvidsson}{2020}]{andersson2020generative}
Filip Andersson and Simon Arvidsson.
\newblock Generative adversarial networks for photo to hayao miyazaki style
  cartoons.
\newblock {\em arXiv preprint arXiv:2005.07702}, 2020.

\bibitem[\protect\citeauthoryear{Bengio \bgroup \em et al.\egroup
  }{2013}]{bengio2013setimating}
Yoshua Bengio, Nicholas L{\'e}onard, and Aaron Courville.
\newblock Estimating or propagating gradients through stochastic neurons for
  conditional computation.
\newblock {\em arXiv preprint arXiv:1308.3432}, 2013.

\bibitem[\protect\citeauthoryear{Chang \bgroup \em et al.\egroup
  }{2020}]{chang2020mix}
Sung-En Chang, Yanyu Li, Mengshu Sun, Runbin Shi, Hayden K-H So, Xuehai Qian,
  Yanzhi Wang, and Xue Lin.
\newblock Mix and match: A novel fpga-centric deep neural network quantization
  framework.
\newblock {\em arXiv preprint arXiv:2012.04240}, 2020.

\bibitem[\protect\citeauthoryear{Chen \bgroup \em et al.\egroup
  }{2021}]{chen2021gans}
Xuxi Chen, Zhenyu Zhang, Yongduo Sui, and Tianlong Chen.
\newblock Gans can play lottery tickets too.
\newblock {\em arXiv preprint arXiv:2106.00134}, 2021.

\bibitem[\protect\citeauthoryear{Chong and Forsyth}{2021}]{chong2021gans}
Min~Jin Chong and David Forsyth.
\newblock Gans n'roses: Stable, controllable, diverse image to image
  translation (works for videos too!).
\newblock {\em arXiv preprint arXiv:2106.06561}, 2021.

\bibitem[\protect\citeauthoryear{Fu \bgroup \em et al.\egroup
  }{2020}]{fu2020autogandistiller}
Yonggan Fu, Wuyang Chen, Haotao Wang, Haoran Li, Yingyan Lin, and Zhangyang
  Wang.
\newblock Autogan-distiller: Searching to compress generative adversarial
  networks.
\newblock In {\em ICML}, 2020.

\bibitem[\protect\citeauthoryear{Guan \bgroup \em et al.\egroup
  }{2020}]{guan2020dais}
Yushuo Guan, Ning Liu, Pengyu Zhao, Zhengping Che, Kaigui Bian, Yanzhi Wang,
  and Jian Tang.
\newblock Dais: Automatic channel pruning via differentiable annealing
  indicator search.
\newblock {\em arXiv preprint arXiv:2011.02166}, 2020.

\bibitem[\protect\citeauthoryear{Han \bgroup \em et al.\egroup
  }{2015}]{han2015learning}
Song Han, Jeff Pool, et~al.
\newblock Learning both weights and connections for efficient neural network.
\newblock In {\em NeurIPS}, 2015.

\bibitem[\protect\citeauthoryear{He \bgroup \em et al.\egroup
  }{2017}]{he2017channel}
Yihui He, Xiangyu Zhang, and Jian Sun.
\newblock Channel pruning for accelerating very deep neural networks.
\newblock In {\em ICCV}, 2017.

\bibitem[\protect\citeauthoryear{Huang \bgroup \em et al.\egroup
  }{2018}]{huang2018munit}
Xun Huang, Ming-Yu Liu, Serge Belongie, and Jan Kautz.
\newblock Multimodal unsupervised image-to-image translation.
\newblock In {\em ECCV}, 2018.

\bibitem[\protect\citeauthoryear{Isola \bgroup \em et al.\egroup
  }{2017}]{isola2018imagetoimage}
Phillip Isola, Jun-Yan Zhu, Tinghui Zhou, and Alexei~A Efros.
\newblock Image-to-image translation with conditional adversarial networks.
\newblock In {\em CVPR}, 2017.

\bibitem[\protect\citeauthoryear{Jin \bgroup \em et al.\egroup
  }{2021}]{jin2021teachers}
Qing Jin, Jian Ren, Oliver~J Woodford, Jiazhuo Wang, Geng Yuan, Yanzhi Wang,
  and Sergey Tulyakov.
\newblock Teachers do more than teach: Compressing image-to-image models.
\newblock In {\em CVPR}, 2021.

\bibitem[\protect\citeauthoryear{Karras \bgroup \em et al.\egroup
  }{2020}]{Karras2019stylegan2}
Tero Karras, Samuli Laine, Miika Aittala, Janne Hellsten, Jaakko Lehtinen, and
  Timo Aila.
\newblock Analyzing and improving the image quality of {StyleGAN}.
\newblock In {\em CVPR}, 2020.

\bibitem[\protect\citeauthoryear{Kim \bgroup \em et al.\egroup
  }{2020}]{Kim2020U-GAT-IT:}
Junho Kim, Minjae Kim, Hyeonwoo Kang, and Kwang~Hee Lee.
\newblock U-gat-it: Unsupervised generative attentional networks with adaptive
  layer-instance normalization for image-to-image translation.
\newblock In {\em ICLR}, 2020.

\bibitem[\protect\citeauthoryear{Lee \bgroup \em et al.\egroup }{2018}]{DRIT}
Hsin-Ying Lee, Hung-Yu Tseng, Jia-Bin Huang, Maneesh~Kumar Singh, and
  Ming-Hsuan Yang.
\newblock Diverse image-to-image translation via disentangled representations.
\newblock In {\em ECCV}, 2018.

\bibitem[\protect\citeauthoryear{Li \bgroup \em et al.\egroup
  }{2020}]{li2020gan}
Muyang Li, Ji~Lin, Yaoyao Ding, Zhijian Liu, Jun-Yan Zhu, and Song Han.
\newblock Gan compression: Efficient architectures for interactive conditional
  gans.
\newblock In {\em CVPR}, 2020.

\bibitem[\protect\citeauthoryear{Li \bgroup \em et al.\egroup
  }{2021}]{li2021anigan}
Bing Li, Yuanlue Zhu, Yitong Wang, Chia-Wen Lin, Bernard Ghanem, and Linlin
  Shen.
\newblock Anigan: Style-guided generative adversarial networks for unsupervised
  anime face generation.
\newblock {\em IEEE TMM}, 2021.

\bibitem[\protect\citeauthoryear{Liu \bgroup \em et al.\egroup
  }{2017}]{liu2018unsupervised}
Ming-Yu Liu, Thomas Breuel, and Jan Kautz.
\newblock Unsupervised image-to-image translation networks.
\newblock {\em NeurIPS}, 2017.

\bibitem[\protect\citeauthoryear{Liu \bgroup \em et al.\egroup
  }{2019}]{liu2019few}
Ming-Yu Liu, Xun Huang, Arun Mallya, Tero Karras, Timo Aila, Jaakko Lehtinen,
  and Jan Kautz.
\newblock Few-shot unsupervised image-to-image translation.
\newblock In {\em ICCV}, 2019.

\bibitem[\protect\citeauthoryear{Park \bgroup \em et al.\egroup
  }{2020}]{park2020cut}
Taesung Park, Alexei~A. Efros, Richard Zhang, and Jun-Yan Zhu.
\newblock Contrastive learning for unpaired image-to-image translation.
\newblock In {\em ECCV}, 2020.

\bibitem[\protect\citeauthoryear{Tang \bgroup \em et al.\egroup
  }{2019}]{tang2019multi}
Hao Tang, Dan Xu, Nicu Sebe, Yanzhi Wang, Jason~J Corso, and Yan Yan.
\newblock Multi-channel attention selection gan with cascaded semantic guidance
  for cross-view image translation.
\newblock In {\em CVPR}, 2019.

\bibitem[\protect\citeauthoryear{Tang \bgroup \em et al.\egroup
  }{2021}]{tang2021attentiongan}
Hao Tang, Hong Liu, Dan Xu, Philip~HS Torr, and Nicu Sebe.
\newblock Attentiongan: Unpaired image-to-image translation using
  attention-guided generative adversarial networks.
\newblock {\em IEEE TNNLS}, 2021.

\bibitem[\protect\citeauthoryear{Wang \bgroup \em et al.\egroup
  }{2020}]{wang2020ganslimming}
Haotao Wang, Shupeng Gui, Haichuan Yang, Ji~Liu, and Zhangyang Wang.
\newblock Gan slimming: All-in-one gan compression by a unified optimization
  framework.
\newblock In {\em ECCV}, 2020.

\bibitem[\protect\citeauthoryear{Yu \bgroup \em et al.\egroup
  }{2017}]{yu2017compressing}
Xiyu Yu, Tongliang Liu, Xinchao Wang, and Dacheng Tao.
\newblock On compressing deep models by low rank and sparse decomposition.
\newblock In {\em CVPR}, 2017.

\bibitem[\protect\citeauthoryear{Zhong \bgroup \em et al.\egroup
  }{2018}]{zhong2018practical}
Zhao Zhong, Junjie Yan, Wei Wu, Jing Shao, and Cheng-Lin Liu.
\newblock Practical block-wise neural network architecture generation.
\newblock In {\em CVPR}, 2018.

\bibitem[\protect\citeauthoryear{Zhu \bgroup \em et al.\egroup
  }{2017}]{CycleGAN2017}
Jun-Yan Zhu, Taesung Park, Phillip Isola, and Alexei~A Efros.
\newblock Unpaired image-to-image translation using cycle-consistent
  adversarial networks.
\newblock In {\em ICCV}, 2017.

\bibitem[\protect\citeauthoryear{Zoph and Le}{2017}]{zoph2016neural}
Barret Zoph and Quoc~V. Le.
\newblock Neural architecture search with reinforcement learning.
\newblock In {\em ICLR}, 2017.

\end{thebibliography}
